# LEADER FOLLOWER FORMATION CONTROL OF GROUND VEHICLES USING CAMSHIFT BASED GUIDANCE

S.M.Vaitheeswaran, Bharath.M.K, and Gokul.M

Aerospace Electronics and Systems Division, CSIR-National Aerospace laboratories, HAL Airport Road, Kodihalli, Bengaluru, 560017 India

## ABSTRACT

*Autonomous ground vehicles have been designed for the purpose of that relies on ranging and bearing information received from forward looking camera on the Formation control . A visual guidance control algorithm is designed where real time image processing is used to provide feedback signals. The vision subsystem and control subsystem work in parallel to accomplish formation control. A proportional navigation and line of sight guidance laws are used to estimate the range and bearing information from the leader vehicle using the vision subsystem. The algorithms for vision detection and localization used here are similar to approaches for many computer vision tasks such as face tracking and detection that are based color-and texture based features, and non-parametric Continuously Adaptive Mean-shift algorithms to keep track of the leader. This is being proposed for the first time in the leader follower framework. The algorithms are simple but effective for real time and provide an alternate approach to traditional based approaches like the Viola Jones algorithm. Further to stabilize the follower to the leader trajectory, the sliding mode controller is used to dynamically track the leader. The performance of the results is demonstrated in simulation and in practical experiments.*

## 1. INTRODUCTION

The problem of controlling a formation of ground vehicles of autonomous ground vehicles continues to draw the attention of researches, thanks to its applications in complex tasks like search and rescue [1], mapping of unknown environments [2], and mobile robotics [3].By formation control we simply mean the problem of controlling the relative position and orientation of the vehicles in a group while allowing the group to move as a whole [4]. The use of formation control enables one to achieve complex tasks with a low robot complexity since each particular task can be specialized in one task of manipulation or robot navigation. Perception, path generation and control characteristics, generally decide the formation control process for their robustness. Also the system architectures provide the infrastructure upon which formation control is implemented and determines the capabilities and limitations of the system. Existing approaches to robot formation control generally fall into three categories: behavior based [5], virtual structure [6] and leader following [7].

In the leader-follower approach, a robot of the formation, designed as the leader, moves along a predefined trajectory while the other robots, the followers, are required to maintain a desired posture (distance and orientation) with respect to the leader [7]. The primary advantage with the approach is that it reduces the formation control problem to a tracking problem where stability of the tracking error can be obtained through standard control theoretic techniques.

One way of realizing formation control in this approach is to shift the domain of the computation space into the vision domain of the follower so that the formation control problem is derived in the local image plane of the follower only undergoing planar motion. The image processing algorithms are then made in real time. using suitable hardware. There has been many attempts to develop fast and simple image analysis methods in real time using a set of visual markers or a set of feature points in real time. The relative pose is then obtained using a geometric pose estimation technique. The fastest object detection in real time is provided by the Viola Jones algorithm using a cascades of classifiers based on Haar-like features . The features is robust to motion blur and changing lighting conditions because those features represent basic elements of the picture including edges, lines and dots. The cascades are trained using AdaBoost learning and calculated with a very fast using integral image Result shows that the system can detect pedestrian at 17 frames per second on 1.7 GHz processor with pedestrian detection rate up to 90% of accuracy [9].

As an alternative, in this paper, we implement for the real time imaging and processing, a color and texture based algorithms developed for vision detection and localization that are also simple but effective and are similar to approaches used for many computer vision tasks such as face tracking and detection.The object detection system is generalized for the formation control problem using vision, and is capable of handling partially occluded vehicles, improves the sensing conditions by accounting for lighting and environmental conditions in addition to real time implementation. In this the traditional categories of algorithms for tracking, namely methods based on target localization and representation are judiciously combined with methods based on filtering and data association.For on-line tracking the Continuously Adapted Mean Shift (CAMShift) [10], [11] algorithm is used which allows on-time tracking. Kalman filters are used to keep track of the object location and predict the location of the object in subsequent frames to help the CAMShift algorithm locate the target in the next. A second filter tracks the leader as it passes under occlusions by using the velocity and position of the object as it becomes occluded to maintain a search region for the CAMShift function as the target reappears from the occlusion. A third filter tracks the area returned by the CAMShift algorithm and monitors changes in area to detect occlusions early.

## 2. PREVIOUS WORK

The purpose of this paper is vision based tracking control in a leader follower formation. It includes occlusion handling, lighting and environmental considerations. Hence we review these state of art detection domains Considerable work has been done in visual tracking to overcome the difficulties arising from noise, occlusion, clutter and changes.

In general tracking algorithms fall into two categories: a) methods based on filtering and data association, b) methods based on target representation and localization.

Tracking algorithms relying on target representation and localization are based on mostly measurements. They employ a model of the object appearance and try to detect this model in consecutive frames of the image sequence. Color or texture features of the object, are used to create a histogram. The object's position is estimated by minimizing a cost function between the model's histogram and candidate histograms in the next image. An example of the method in this category is the mean shift algorithm where the object inside an ellipse and the histogram is constructed from pixel values inside that ellipse [12]. The extensions of the main algorithm are proposed in [13]. The mean shift is combined with particle filters in [14]. Scale invariant features are used in [15] where various distance measures are associated with the mean shift algorithm. These methods have the drawback that the type of object's movement should be correctly modeled.

The algorithms based on filtering assume that the moving object has an internal state which may be measured and, by combining the measurements with the model of state evolution, the object's position is estimated. An example of this category is the Kalman filter [16] which successfully tracks objects even in the case of occlusion [17], the particle filters [13,14] the Condensation [18] and ICondensation [19] algorithms which are more general than Kalman filters These algorithms. have the ability to predict an object's location under occlusion as well and use factored sampling.

## 3. PRESENTED APPROACH FOR TRACKING USING THE VISION SYSTEM

The presented work combines in real time the advantages of pose representation through color and texture, localization through the CAMShift algorithm and uses the filtering and data association problem through Kalman filtering for occlusion detection The visual tracking problem is divided into target detection and pose estimation problem The overall flow diagram of the proposed approach for target detection is described in Fig.1. The color histogram matching is used to get a robust target identification which is fed to CAMShift algorithm for robust tracking. In a continuous incoming video sequence, the change of the location of the object leads to dynamic changes of the probability distribution of the target. CAMShift changes the probability distribution dynamically, and adjusts the size and location of the searching window based on the change of probability distribution. The back projected image is given as input for CAMShift processing for tracking the target. After each frame is read, the target histogram is updated to account for any illumination or color shift changes of the target. This is done by computing a candidate histogram of the target in the same manner the original target histogram is computed. Then the candidate and target histograms are compared using the histogram comparison technique that uses a Bhattacharyya Coefficient [20]. The Bhattacharyya Coefficient returns a value of 0 or 1, with 0 being a perfect match and 1 being a total mismatch of the histograms. Once a region of interest (ROI) is measured, the algorithm creates a vector that shifts the focus of the tracker to the new centre of density.

### 3.1. Object representation using color

Color is used in identifying and isolating objects. The RGB values of every pixel in the frame are read and converted into the HSL color space. The HSL color space is chosen because the chromatic information is independent from the lighting conditions. Hue specifies the base color, saturation determines the intensity of the color and luminance is dependent on the lighting condition. Since every color has its own range of H values, the program compares the H values of every pixel with a predefined range of H values of the landing zone. If it falls within 10%, the pixel is marked out as being part of the landing zone. With the range of H correctly chosen, the landing spot is identified more accurately. After going through all the pixels for one frame, the centroid of all the marked pixels is calculated. From this result, the relative position of the landing zone with respect to the center of the camera's view is known. The conversion from RGB color space to HSV color space is performed using equation (1).Here red (r), green (g), blue (b) $\in[0,1]$ are the coordinates of the RGB color space and  max and minicorrespond to the greatest and least of r, g and b respectively .The Hue angle h$\in$[0,360] for HSV color space  is given by [21].

$$h = \begin{cases} 0, & if\ \max = \min, \\ \left(60^{\mathrm{O}} * \dfrac{g-b}{\max-\min} + 360^{\mathrm{O}}\right), & if\ \max = r, \\ 60^{\mathrm{O}} * \dfrac{b-r}{\max-\min} + 120^{\mathrm{O}}, & if\ \max = g, \\ 60^{\mathrm{O}} * \dfrac{r-g}{\max-\min} + 240^{\mathrm{O}}, & if\ \max = b \end{cases} \quad (1)$$

The value of h is normalized to lie between 0 and $180^0$ to fit into an 8 bit gray scale image (0-255), and h= 0 is used when max=min, though the hue has no geometric meaning for gray. The s and v values for HSV color space are defined as follows:

$$s = \begin{cases} 0, & if\ \max = 0, \\ \dfrac{\max-\min}{\max} = 1 - \dfrac{\min}{\max}, & otherwise \end{cases} \quad (2)$$

The v or value channel represents the gray scale portion of the image. A threshold for the Hue value of the image is set based on the mounted marker color. Using the threshold value, segmentation between the desired color information and other colors is performed. The resulting image is a binary image with white indicating the desired color region ad black assumed to be the noisy region. The contour of desired region is obtained as described in the section below.

### 3.2 Contour detection

Freeman chain code [22] method is used for finding contours, which is based on 8 connectivity of 3x3 windows of Freeman chain code. Two factors determine the success of the algorithm: The first factor is the direction of traverses either clockwise or anticlockwise. The other is the start location of the 3X3 window traverse. Chain code scheme is a representation that consists of series of numbers which represent the direction of the next pixel that can be used to represent shape of the objects. Chain code is a list of codes ranging from 0 to 7 in clockwise direction representing the direction of the next pixel connected in 3X3 windows as shown in Fig. 2. The coordinate of the next pixel is calculated based on the addition and subtraction of columns and rows by 1, depending on the value of the chain code.

### 3.3 Background projection

A back projection image is obtained using the Hue, Saturation and local binary pattern channels along with the target histogram. The back projection image is a mono channel image whose pixel value probability range between 0to255 that corresponds to the probability of the pixel values in the ROI.

The histogram process categorizes the value of each pixel in the selected region and assigns each into one of N bins, corresponding to N bins of the histogram dimension. In this case a three-dimensional histogram is used with dimension of 32 (hue) X 16 (saturation) x 36 (LBP) bins = 18,432 bins. In a similar way, a histogram is created for the remainder of the background to identify the predominant values not in the target. Weights are assigned for each of the target bins such that the target values unique to the target will have a higher relative value versus hues that are in common with the background. The resulting histogram lookup table maps the input image plane value to the bin count, and then normalizes from 0 to 255 to ensure the resulting grey scale image has valid pixel intensity values. In the paper, the initial target histogram lookup table is created at target selection and is saved as a reference histogram for later processing. The latest target histogram is then used for each frame to map the hue, saturation image planes into a resulting back projection image used by the CAMShift process. The resulting image from this process has mostly white pixels (255) at the predominant target locations and lower values for the remaining colors.

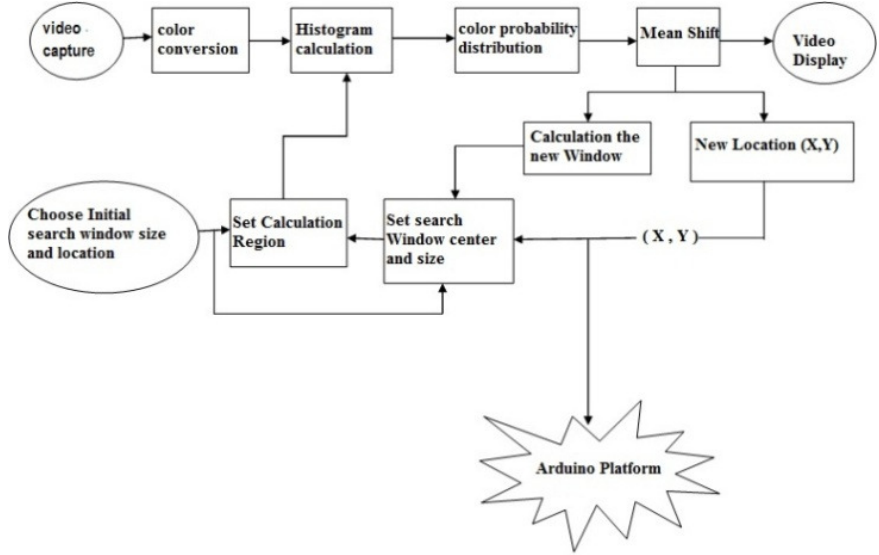

Figure 1. Flowchart of the Vision Algorithm

| 5 | 6 | 7 |
|---|---|---|
| 4 | Pixel Position X,Y | 0 |
| 3 | 2 | 1 |

| Current pixel position x, y | | |
|---|---|---|
| Code | Next row | Next column |
| 0 | X | y+1 |
| 1 | x-1 | y+1 |
| 2 | x-1 | Y |
| 3 | x-1 | y-1 |
| 4 | X | y-1 |
| 5 | x+1 | y-1 |
| 6 | x+1 | Y |
| 7 | x+1 | y+1 |

Figure 2. Freeman Chain code representation

## 3.4 Tracking using CAMShift

The inputs to the CAMShift algorithm are window coordinates to restrict the search, and a back projection image where the pixel values represent the relative probability that the pixel contains the hues, saturation. CAMShift algorithm operates on the probability distribution image that is derived from the histogram of the object to be tracked generate above.

The principle steps of the CAMShift algorithm are as follows:

1) Choose the initial location of the mean shift search window.
2) Calculate the 2D color histogram within the search window.
3) Perform back-projection of the histogram to a region of interest (ROI) centered at the search window but slightly larger the mean shift window size.

25

4) Iterate Mean Shift algorithm to find the centroid of the probability image and store the zeroth moment and centroid location. The mean location within the search window of the discrete probability

$$M_{0,0} = \sum_x \sum_y I(x, y) \qquad (3)$$

image is found using moments. Given that I(x, y) is the intensity of the discrete probability image at (x, y) within the search window, the zeroth moment is computed as:(3)
The first Moment for x and y is,

$$M_{0,1} = \sum_x \sum_y yI(x, y)$$
$$M_{1,0} = \sum_x \sum_y xI(x, y) \qquad (4)$$

Then the mean search window location can be found as:

$$(x_c, y_c) = \left( \frac{M_{10}}{M_{00}}, \frac{M_{01}}{M_{00}} \right) \qquad (5)$$

For the next video frame, center the search window at the mean location stored in Step 4 and set the window's size to a function of the zero$^{th}$ moment. Go to Step 2. The scale of the target is determined by finding an equivalent rectangle that has the same moments as those measured from the probability distribution image. Define the second moments as:

$$M_{2,0} = \sum_x \sum_y x^2 I(x, y)$$
$$M_{0,2} = \sum_x \sum_y y^2 I(x, y) \qquad (6)$$
$$M_{1,1} = \sum_x \sum_y xyI(x, y)$$

The following intermediate variables are used

$$\left. \begin{array}{l} a = \dfrac{M_{2,0}}{M_{0,0}} - x_c^2 \\[6pt] b = \dfrac{M_{1,1}}{M_{0,0}} - xy \\[6pt] c = \dfrac{M_{0,2}}{M_{0,0}} - y \end{array} \right\} \qquad (7)$$

Then the dimension of the search window can be computed as:

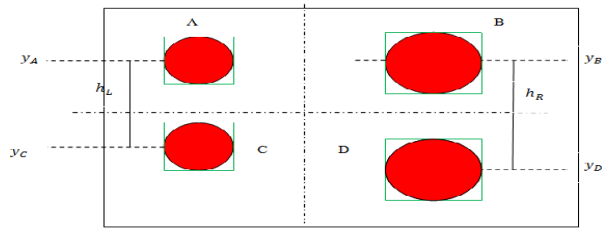

Figure 3. Detected pattern on Follower Image

$$h = \sqrt{\frac{(a+c) - \sqrt{b^2 + (a-c)^2}}{2}}$$

$$w = \sqrt{\frac{(a+c) + \sqrt{b^2 + (a-c)^2}}{2}}$$

(8)

## 3.5 Pose Estimation

The image measurements from the follower are the relative distance and the relative bearing in the leader follower framework. The follower vehicle is equipped with a forward facing camera mounted on it. Camera pose calculation from one frame to the other requires information about objects that are viewable in the frames. Information for this is obtained from the feature extraction method proposed above. The camera captures features from the pattern mounted on the leader robot, as shown in Fig.3.The four markers, are converted to high intensity values in the HSV channels and localized from frame to frame using the CAMShift-Back ground projection filtering process.

From these feature positions, the relative posture of the follower with respect to leader is determined. The equations are

$$x_T = \frac{x_R + x_L}{2}, z_T = \frac{z_R + z_L}{2}, \varphi_T = \cos^{-1}\left(\frac{x_R - x_L}{E}\right)$$

(9)

Where E is the length of the sides of the square. Using inverse perspective projection the measurement equations is formulated as shown in Fig. 4.

$$x_L = \frac{z_L - f}{f} x_A, x_R = \frac{z_R - f}{f} x_B$$

$$z_L = f\left\{\frac{E}{h_L} + 1\right\}, z_R = f\left\{\frac{E}{h_R} + 1\right\}$$

(10)

Using variables of equations (6) and (7), and assuming the robot centre and camera centre coincide, the relative position and angular displacement with respect to the target vehicle ($\varphi$, D) can be calculated using the following equations.

Figure 4. Target position with respect to camera centre

The International Journal of Multimedia & Its Applications (IJMA) Vol.6, No.6, December 2014$$\left. \begin{array}{l} \varphi = \tan^{-1}\left(\dfrac{x_T}{Z_T}\right) \\ \phi = \tan^{-1}\left(\dfrac{Z_R - Z_L}{X_R - X_L}\right) \\ \theta = \varphi + \phi \\ D = \sqrt{(x_T^2 + Z_T^2)} \end{array} \right\} \quad (11)$$

### 3.6 Vehicle Control

In our case, we have two robots; the tasks of each robot are different and based on their role. Leader has as primary task to track a pre-defined trajectory using a robust controller based on sliding mode control [23]. Follower uses a visual feedback and tries to identify and track the trajectory of the leader. Control laws for the follower vehicle are based on incorporating range and bearing to the leader vehicle that is received from the camera. Based on the range between the leader and the follower vehicle, and line of sight guidance definitions as shown in [24] with reference to Fig. 4 the follower vehicle adjusts its speed (U) in order to achieve the command range

As shown in Fig.5, (X(i-1), Y(i-1)), (X(i), Y(i)) and (X(i+1), Y(i+1)) are the ((i-1)[th], i[th] and (i+1)[th] waypoints (denoted as 'o') respectively assigned by the mission plan. The LOS guidance employs the line of sight between the vehicle and the targets The LOS angle to the next waypoint is

defined as $\psi_{los} = \arctan\dfrac{Y_i - y(t)}{X_i - x(t)}$ (12)

where inertial positions are $((x(t), y(t), \psi_d(i))$.

The angle of current line of tracking is $\psi_d(i) = \arctan\dfrac{Y_i - Y_{i-1}}{X_i - X_{(i-1)}}$ (13)

The cross track error is given by:
$$\tilde{\psi}(t)_{crosstracker\,ror(i)} = \psi_d(i) - \psi(t) \quad (14)$$
The distance from current to the next waypoint is:
$$S(t)_i = \sqrt{\tilde{X}(t)_i^2 + \tilde{Y}(t)_i^2} \quad (15)$$
Where the cross track error is given by:
$$\begin{array}{l} \varepsilon(t) = S(t)_i \sin(d_p(t)) \\ d_p(t) = \psi_{los} - \psi_d(i) \end{array} \quad (16)$$
Range error is defined as:
$$\tilde{S} = S(t) - S_{com} \quad (17)$$

To determine the speed of the follower vehicle, the following formulas are defined [24]:
$$\dot{\sigma}_z = -\eta\,\mathrm{sgn}(\sigma_z) \quad (18)$$

28

where $\sigma_z = \tilde{z}$ Also,

$$\dot{\sigma}_z = \dot{\tilde{z}} = (\dot{S} - 0) \qquad (19)$$

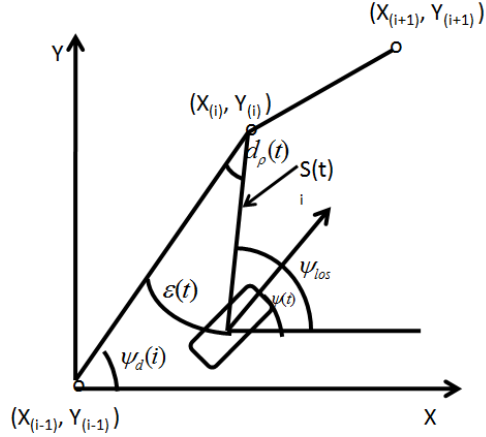

Figure 5. Track Geometry used for Line of Sight Guidance

Where $\dot{S}$ is defined by

$$\dot{S} = \left(\frac{x_0 - x}{z}\right)(\dot{x}_0 - \dot{x}) + \left(\frac{y_0 - y}{z}\right)(\dot{y}_0 - \dot{y}) \qquad (20)$$

Where,
$x_0$ is the leader's x-position
$y_0$ is the leader's y-position
x is the follower's x-position
y is the follower's y-position

$$\dot{x}_0 = U_0 \sin(\psi_0) \qquad (21)$$

$$\dot{x} = U \sin(\psi) \qquad (22)$$

Combining Equations (21) and (22) and solving for U results in the following formula:

$$U = \frac{-\text{sgn}(\tilde{z})z + \dot{x}_0(x - x_0) + \dot{y}_0(y - y_0)}{\cos\psi(x - x_0) + \sin\psi(y - y_0)} \qquad (23)$$

The follower vehicle adjusts its heading ($\psi_f$) in order to achieve the command bearing ($\beta\, com$) depending upon the bearing between the leader and the follower vehicle. Bearing error is defined as:

$$\tilde{\beta} = (\beta(t) - \beta_{com}) \qquad (24)$$

In order to determine the heading of the follower vehicle the following formulas are defined:

$$\dot{\sigma}_\beta = -\eta\, \text{sgn}\left(\tilde{\beta} + \lambda \dot{\tilde{\beta}}\right) \qquad (25)$$

$$\delta = -\eta\, \text{sgn}\left(\tilde{\beta} + \lambda r\right) \qquad (26)$$

## 4. RESULTS

To test the effectiveness of the vision algorithms, experiments have been consisting of two ground vehicles in a convoy formation using autonomous control for the master and vision control for the slave robot which uses a low cost vision sensor(camera) as the only sensor to obtain the location information. The robot vehicle frame and chassis is a model manufactured by Robokit. It has a differential 4-wheeled robot base with two wheels at the front and the two wheels are at the back. The vehicle uses four DC motors (geared) for driving the four wheels independently. The DC motors are controlled by a microcontroller. For the vision algorihm the coding is performed in python language with OpenCV library modules ported to python block. Practical experiments were conducted in order to test the performance and robustness of the proposed vision based target following which are described in the forth coming sections.

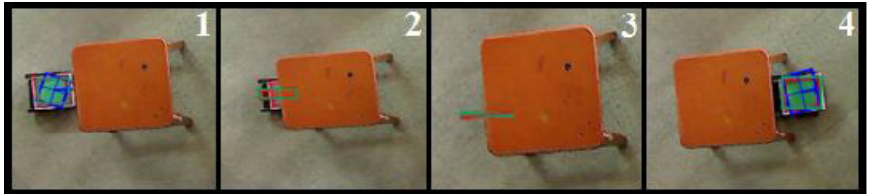

Figure 6. Occlusion detection and recovery of target vehicle in real time

### 4.1 Partially and fully occluded target detection

Figure 6 illustrates the case where the target is partially (sequence 2) and fully occluded (sequence 3) while moving through an occlusion (wooden stool) .The recovery from occlusion is illustrated in Fig. 6 (sequence 4). This process was repeated for multiple environments, in real-time, and the algorithm works robustly for all of the used tests, showing the overall effectiveness of this algorithm. The recovery was found to be accurate in all cases experimentally studied. Target Detection in presence of same/similar color and texture Fig. 8 demonstrates the efficacy of the detection process using the traditional CAMShift using color based features.
.

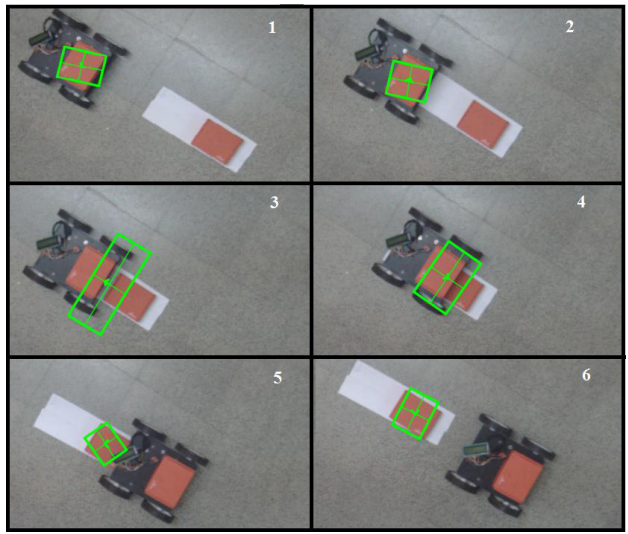

Figure 7: Target tracking with similar colored object using traditional CAMShift

The International Journal of Multimedia & Its Applications (IJMA) Vol.6, No.6, December 2014

As can be seen from the video sequence of Fig. 8, because the hues are similar and are of similar texture to the tracked markers of the leader vehicle, the tracker leaves the banner (marker on the leader vehicle) and remains locked to the background object. Fig.9 shows the improvement to the detection process obtained by adding the saturation and texture functions of the leader vehicle in the presence of similar color/textured objects in the background. The hue + saturation + texture method works very well in tracking the leader through the background that is of a nearly identical color as the leader red marker. There is enough difference in texture that the back projection algorithm is able to create a probability density function that correctly isolates the leader from the background

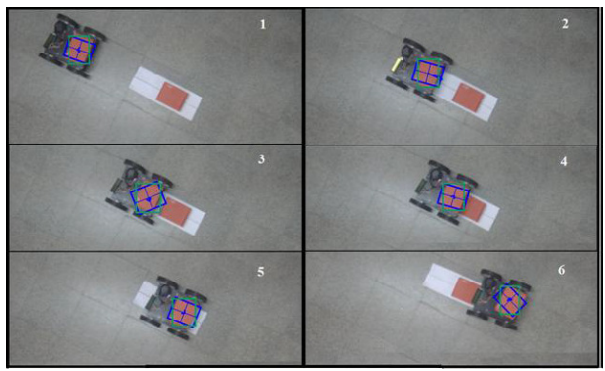

Figure 8. Target tracking with similar colored background using proposed CAMShift+ LBP + Kalman

## 4.2 Illumination Variation

Fig. 10 highlights the improvements made by the addition of saturation and texture channels to the back projection process when compared to the CAMShift algorithm using hue only (Figure 9) The improvement on how the leader is isolated from the similar illuminated background is clearly evident.

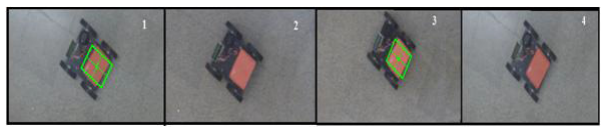

Figure 9. Target tracking response with indoor light variation using CAMShift

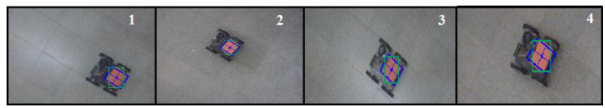

Figure 10. Target tracking response with indoor light variation using CAMShift+LBP+Kalman

## 5. CONCLUSIONS

The algorithms developed for vision detection and localization provide a simple but effective processing approach in real time for the leader-follower formation control approach considered in the paper. The generalized framework proposed using the Vision Algorithm presented based on Hue+ Saturation + texture when combined with Kalman filters accounts for full and partial occlusion of the leader, the lighting and environmental conditions and overcomes practical limitations that exist in the information gathering process in the leader follower framework. The CAMShift tracker based on hue only becomes confused when the background and the object have similar hues, as it is unable to differentiate the two. Selection of appropriate Saturation and Luminance thresholds and implementation in a weighted histogram reduces the impact of common hues and enhances the performance detections for the CAMShift tracker in the leader follower framework. The addition of Local Binary Patterns greatly increases track performance. The addition of the dynamic updating of the target histogram to accounts for lighting improves the track for long sequences or sequences with varying lighting conditions. The controller using line of sight guidance for the follower faithfully approaches the leader by maintaining the speed and distance as demanded, indicating the validity of the approach and method used.